\documentclass[conference]{IEEEtran}
\IEEEoverridecommandlockouts
\usepackage{cite}
\usepackage{amsmath,amssymb,amsfonts}
\usepackage{algorithmic}
\usepackage{imakeidx}
\usepackage{booktabs}
\usepackage{graphicx}
\usepackage{multirow}
\usepackage{float}
\usepackage{subfigure}
\usepackage{bbding}
\usepackage{textcomp}
\usepackage{xcolor}
\usepackage{hyperref}
\usepackage{tabularx}
\usepackage{comment}
\usepackage[capitalise,noabbrev]{cleveref}
\def\BibTeX{{\rm B\kern-.05em{\sc i\kern-.025em b}\kern-.08em
    T\kern-.1667em\lower.7ex\hbox{E}\kern-.125emX}}
\usepackage{geometry}
\begin{document}
\newgeometry{left=0.75in,right=0.75in,top=1in,bottom=0.75in}

\title{UniSTPA: A Safety Analysis Framework for End-to-End Autonomous Driving}

\author{%
  Hongrui Kou$^{1,*}$, 
  Zhouhang Lyu$^{1,*}$, 
  Ziyu Wang$^{1}$, 
  Cheng Wang$^{2}$, 
  Yuxin Zhang$^{1,\dagger}$%
    \thanks{This work was supported in part by the Science and Technology Development Program of Jilin Province under Grant 20240302052GX and in part by the National Natural Science Foundation of China under Grant 52075213.} 
  \thanks{$^{1}$National Key Laboratory of Automotive Chassis Integration and Bionics, Jilin University, Changchun, China. {\tt\small\{kouhr23, lvzh22, zyw22, yuxinzhang\}@jlu.edu.cn.}}%
  \thanks{$^{2}$School of Engineering and Physical Sciences, Heriot‐Watt University, Edinburgh, United Kingdom. {\tt\small Cheng.Wang@hw.ac.uk.}}%
  \thanks{* Co-first authors.}%
  \thanks{† Corresponding author.}%
}

\maketitle

\begin{abstract}
As autonomous driving technology continues to advance, end-to-end models have attracted considerable attention owing to their superior generalisation capability. Nevertheless, such learning-based systems entail numerous safety risks throughout development and on-road deployment, and existing safety-analysis methods struggle to identify these risks comprehensively. To address this gap, we propose the \emph{Unified System Theoretic Process Analysis} (UniSTPA) framework, which extends the scope of STPA from the operational phase to the entire lifecycle of an end-to-end autonomous driving system, including information gathering, data preparation, closed loop training, verification, and deployment. UniSTPA performs hazard analysis not only at the component level but also within the model’s internal layers, thereby enabling fine-grained assessment of inter and intra module interactions. Using a highway Navigate on Autopilot function as a case study, UniSTPA uncovers multi-stage hazards overlooked by conventional approaches including scene design defects, sensor fusion biases, and internal model flaws, through multi-level causal analysis, traces these hazards to deeper issues such as data quality, network architecture, and optimisation objectives. The analysis result are used to construct a safety monitoring and safety response mechanism that supports continuous improvement from hazard identification to system optimisation. The proposed framework thus offers both theoretical and practical guidance for the safe development and deployment of end-to-end autonomous driving systems.
\end{abstract}


\section{Introduction}\label{AA}
With the rapid advancement of intelligent and connected vehicle technology, end-to-end (E2E) autonomous driving has emerged as a novel paradigm and is moving from academic research toward real-world deployment. Contemporary E2E approaches fall into two principal categories \cite{chen2024end}: \emph{i) monolithic} end-to-end, in which the complete perception–planning–control chain is collapsed into a single unified model trained to minimise the loss from sensor inputs to actuator outputs; \emph{ii) modular} end-to-end, which retains functional sub-modules connected through explicit interfaces yet trains them jointly via E2E back-propagation. This study focuses on the \emph{modular} end-to-end paradigm. However, the “black-box” nature of E2E models \cite{le2022survey} poses unprecedented challenges for safety analysis: these models operate with limited internal explainability and structural opacity, while traditional methods typically require clear visibility into component functions and interaction processes. And E2E models have unique safety risks at every stage from development to deployment, necessitating a comprehensive, lifecycle-oriented safety analysis framework.


Safety analysis plays a critical role in the development of autonomous driving systems (ADS), directly impacting functional safety, intended functionality, and information security. Widely adopted in engineering, fault tree analysis (FTA) \cite{ruijters2015fault}, failure modes and effects analysis (FMEA)\cite{liu2013risk}, and hazard and operability analysis (HAZOP) \cite{rossing2010functional} are primarily based on component-failure assumptions. When combined with ISO~26262 \cite{ISO26262}, these methods provide a systematic safety assurance framework for conventional automotive electronic and electrical systems. However, as the level of autonomy increases, system complexity and interactions grow significantly, and these traditional approaches—focusing on deterministic faults—struggle to address hazards originating from complex interactions or algorithmic decision errors \cite{madala2023adsa}.

\begin{table*}[htbp]
  \centering
  \caption{Comparison of Safety Analysis Methods}
  \label{tab:method_comparison}
  \begin{tabularx}{\linewidth}{>{\centering\arraybackslash}l|*{5}{>{\centering\arraybackslash}X}}
    \toprule
    \textbf{Characteristic} & \textbf{FTA} & \textbf{FMEA} & \textbf{HAZOP} & \textbf{STPA} & \textbf{UniSTPA} \\
    \midrule
    Analytical perspective                & Deductive               & Inductive             & Guide-word based         & Systems-theoretic                     & Systems-theoretic        \\
    Component interaction capability      & Weak                    & Weak                  & Moderate                  & Strong                                & Strong                                \\
    Human–machine interaction analysis    & Weak                    & Weak                  & Moderate                  & Strong                                & Strong                                \\
    Applicability to software systems     & Weak                    & Moderate              & Moderate                  & Strong                                & Strong                                \\
    Non-fault-based hazard identification & Weak                    & Weak                  & Moderate                  & Strong                                & Strong                                \\
    Scalability to system complexity      & Limited                 & Limited               & Moderate                  & Good                                  & Excellent                             \\
    Analysis cost and time                & Low                     & Low                   & High                      & High                                  & High                                  \\
    Applicable development phases         & Later phases            & Mid–later phases      & Early phases              & Later phases                & Entire lifecycle                      \\
    Learning-based system analysis & Weak                    & Weak                  & Weak                      & Moderate                              & Strong                                \\
    \bottomrule
  \end{tabularx}
\end{table*}

Systems Theoretic Process Analysis (STPA) \cite{sadeghi2023proposed} as an emerging systems-theoretic safety analysis method, treats the system as a control loop and analyzes unsafe control actions (UCA) and their causes at the system level. By transcending the limitations of component-failure assumptions, STPA can address software bugs, component interaction failures, and human factors. This capability aligns closely with ISO~21448 \cite{ISO21448}, which extends safety analysis to non-fault-based hazards arising from performance limitations and misuse. Nevertheless, existing STPA applications focus primarily on runtime risk control and there is still insufficient analysis of E2E autonomous driving systems that cover the entire life cycle. In particular, for learning-based E2E models, traditional STPA treats the entire model as a single control entity, obscuring hazards at individual functional layers \cite{utesch2020towards}. Although recent standards such as ISO/PAS 8800 \cite{ISO8800} emphasize AI systems’ lifecycle safety management, a dedicated framework for E2E autonomous driving remains lacking.

To address the above issues, this paper proposes a Unified System Theoretic Process Analysis (UniSTPA) framework, which extends the conventional STPA along two dimensions: the \textit{lifecycle} dimension and the \textit{structural-depth} dimension. UniSTPA expands the scope of analysis from a single runtime phase to the system’s full lifecycle and performs safety analysis deep within the internal components of end-to-end models. This paper applies the framework to a Level 3 highway Navigate on Autopilot (NOA) function to identify potential hazards, design safety monitoring and response mechanisms, and provide theoretical and practical guidance for the safe development of E2E autonomous driving systems. The contributions of this work are as follows:
\begin{enumerate}
  \item Introduce the UniSTPA framework, which extends the STPA methodology vertically across the five development and operational phases of an E2E autonomous driving system—information gathering, data preparation, model training, verification, and deployment. It is the first time to systematically identify and control potential hazards of the E2E ADS through safety analysis methods.
  \item Propose a granular control‐structure modeling method for E2E models, incorporating internal modules such as perception, navigation, prediction, and planning into the analysis framework to identify unsafe control actions within and between modules, thereby effectively “decomposing” the AI “black‐box”.

  \item We develop a closed‐loop safety management system that integrates safety monitors and hierarchical response mechanisms, feeding operationally detected risks back into the design and validation processes to establish a continuous safety‐improvement loop across development, verification, and operation.
\end{enumerate}

\section{Relate Work}\label{BB}
\subsection{Safety Analysis Methods}
With the advancement of autonomous driving technologies, the effectiveness and applicability of system safety analysis methods have become focal points of research \cite{qi2023stpa}. These methods can be classified into two major categories according to their theoretical basis and analytical approach: \textit{traditional} methods and \textit{systems‐theoretic} methods. Traditional safety analysis methods rely primarily on event‐chain theory or component‐failure theory, whereas systems‐theoretic methods adopt a holistic system perspective, emphasizing the interactions and constraint relationships among system components \cite{mahajan2017applicationa}. Existing safety analysis methods exhibit three key limitations when applied to E2E autonomous driving systems: \textit{i)} Traditional approaches (FTA, FMEA, HAZOP) rely on component-failure assumptions and cannot adequately analyze hazards arising from component interactions and emergent system behavior\cite{leveson2016engineering}; \textit{ii)} Even systems-theoretic methods such as STPA are predominantly applied to the post-deployment operational phase and lack a systematic, full-lifecycle analysis encompassing data acquisition, model training, and other development stages \cite{abdulkhaleq2017using}; \textit{iii)} All current methods treat deep learning models as monolithic “black-boxes,” rendering them unable to analyze internal module interactions and thus limiting their ability to identify hazards unique to learning-based systems \cite{kuznietsov2024explainable}. 

Table \ref{tab:method_comparison} provides a comparative overview of the principal safety analysis methods. As indicated in Table \ref{tab:method_comparison}, STPA exhibits marked advantages over traditional methods in analyzing complex system interactions and design deficiencies. The UniSTPA framework proposed herein further extends STPA’s analytical scope to encompass the full lifecycle of E2E autonomous driving systems and to enable fine‐grained analysis of the model’s internal structure.

\subsection{Application of STPA in Autonomous Driving}
In recent years, the application of STPA in autonomous driving has increased. In the domain of functional safety, Abdulkhaleq \cite{abdulkhaleq2017using} employed STPA for comprehensive safety analysis of fully autonomous vehicles and concluded that STPA is an effective tool for deriving safety constraints. In the field of Safety of the Intended Functionality (SOTIF), Khastgir et al \cite{khastgir2021systems} proposed an extended method based on STPA for generating test scenarios of automated driving systems, identifying hazards and unsafe control actions, determining scenario elements, and evaluating system safety.
As cybersecurity risks have increased, STPA has also been applied to information security. Young and Leveson \cite{young2013systems}  introduced the STPA‐sec method to incorporate information security issues into the safety analysis process. Schmittner \cite{schmittner2016limitation}  enhanced STPA‐sec and applied it to the safety analysis of hybrid‐vehicle battery management systems. Liu et al \cite{liu2025quantitative} proposed a method combining improved STPA-SafeSec and Bayesian network, extending STPA-SafeSec with the STRIDE threat model for the quantitative risk assessment of connected automated vehicles.
In the domain of human–machine interaction, STPA is well suited to analyzing interaction risks between autonomous driving systems and drivers. France \cite{france2017engineering} proposed STPA‐Engineering for Humans, a human‐controller model for analyzing human–machine interactions and identifying hazard scenarios related to human behavior. Liao et al \cite{liao2025human} proposed a method integrating STPA-IDAC and BN-SLIM, using STPA to determine performance shaping factors causing human errors for the human reliability analysis in the take-over process of Level 3 automated driving.


\begin{figure}[!b]
    \centering
    \includegraphics[width=0.7\linewidth,keepaspectratio]{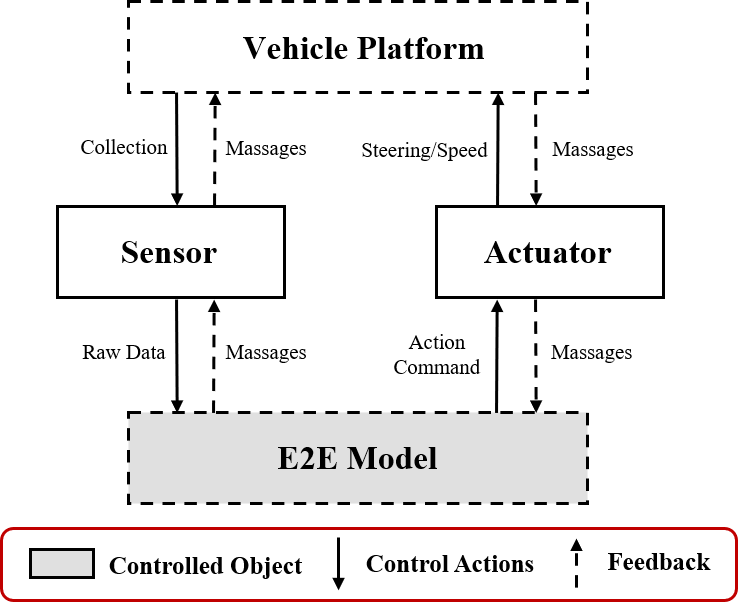}
arxiv
\caption{Traditional STPA Control Structure Applied to E2E ADS.}
    \label{fig: baseSTPA}
\end{figure}

However, existing studies have primarily focused on specific functional modules of conventional autonomous driving systems and lack comprehensive analysis of end‐to‐end autonomous driving systems. 

\section{UniSTPA}\label{CC}
Traditional STPA employs the simplified control structure shown in \hyperref[fig: baseSTPA]{Fig 1}, treating the entire end-to-end autonomous driving system as a single “black-box” controller. Although this model can broadly describe system operation, it has three key limitations: (1) it cannot analyze interactions among the model’s internal modules; (2) it focuses exclusively on the post-deployment operational phase, neglecting potential risks in early development stages such as data acquisition and training optimization; (3) it does not account for attributes unique to learning-based systems, making it difficult to identify hazards arising from data bias, model robustness issues, or inappropriate optimization objectives. These limitations constrain the effectiveness of traditional STPA in analyzing deep learning–based autonomous driving systems.

\begin{figure}[!h]
    \centering
    \includegraphics[width=0.7\linewidth,keepaspectratio]{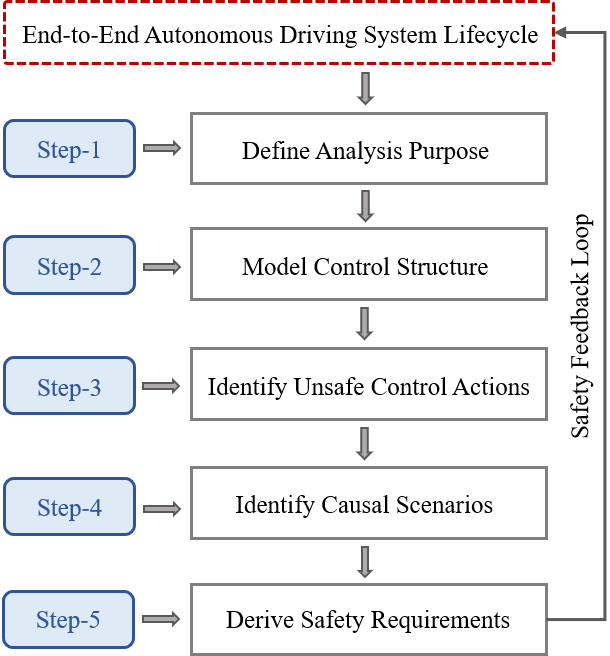}
    \caption{UniSTPA Analysis Framework for E2E ADS.}
    \label{fig: step}
\end{figure}

To overcome these shortcomings, this paper proposes the UniSTPA method, whose core workflow is illustrated in \hyperref[fig: step]{Fig 2}. UniSTPA proceeds through the following steps: Step 1 defines specific hazards for each lifecycle stage; Step 2 constructs a multi-layer control structure spanning the entire process; Step 3 identifies unsafe control actions in each stage and module; Step 4 analyzes multidimensional causal scenarios; Step 5 derives systematic safety requirements (SR). A safety feedback loop integrates the entire analysis process, achieving closed-loop management from requirement specification to verification. This end-to-end, multi-level, full-lifecycle analysis enables UniSTPA to comprehensively identify and manage potential hazards throughout the development and operation of end-to-end autonomous driving systems, thus providing a more robust safety assurance framework.

\section{Case Study}\label{DDD}
This section focuses on the NOA for Highway system equipped with an end-to-end model. The vehicle is equipped with cameras, LiDAR, and other sensors to acquire environmental information, leverages the end-to-end model to generate control commands for steering, acceleration, and braking, thereby achieving both longitudinal and lateral control. We apply the UniSTPA framework to conduct a full-lifecycle safety analysis, illustrating how to identify potential hazards at each stage of the system and derive systematic safety requirements. The case study encompasses STPA hazard identification and causal analysis and further designs dedicated safety monitoring and response mechanisms to enable closed-loop management of operational risks.  
\subsection{Define Analysis Purpose}
The main content of defining analysis purposes include defining losses and identifying system‐level hazards. In the UniSTPA analysis framework, losses denote undesirable outcomes that may occur during system operation, such as personal injury or property damage. Hazards refer to the sets of system states or conditions that, under worst‐case environmental circumstances, can lead to those losses. For the E2E‐based NOA system, we define the loss list as shown in Table \ref{tab:losses}. Although L3 and L4 are not safety‐critical losses, their inclusion enables more comprehensive safety risk management and aligns with the considerations of ISO~21448 regarding system performance limitations and user‐experience–related risks. Traffic violations may not only incur legal liability but also serve as precursors to severe accidents; a reduction in user trust directly affects technology adoption rates and may lead to misuse. 

\begin{table}[htbp]
  \caption{List of Losses}
  \label{tab:losses}
  \centering
  \begin{tabularx}{\columnwidth}{@{}c X@{}}
    \toprule
    \textbf{Loss ID} & \textbf{Description} \\
    \midrule
    L1 & Bodily injury to the vehicle operator, passengers, or other road users. \\
    L2 & Physical damage to the ego vehicle resulting from collisions with other vehicles or static obstacles. \\
    L3 & Traffic violations and operational disruptions, including regulatory infringements, reduced roadway capacity, congestion, or diminished transportation efficiency. \\
    L4 & Erosion of user trust due to inadequate or unpredictable system performance. \\
    \bottomrule
  \end{tabularx}
\end{table}

Based on the losses defined above, Table \ref{tab:hazards} presents the system‐level hazards of the urban NOA system. It comprehensively covers six core safety‐critical scenarios—lane positioning, traffic interaction, decision and trajectory planning, operational boundary violations, traffic‐rule compliance, and degraded‐mode takeover—each of which is associated with one or more system losses, thereby forming a hazard–loss causal chain. This mapping relationship provides a clear basis for the subsequent derivation of safety requirements and the design of hazard control measures.

\begin{table}[!h]
  \caption{System-Level Hazards and Corresponding Losses}
  \label{tab:hazards}
  \centering
  \begin{tabularx}{\columnwidth}{@{}c X X@{}}
    \toprule
    \textbf{Hazard ID} & \textbf{System-Level Hazard} & \textbf{Associated Losses} \\
    \midrule
    H1 & Vehicle fails to keep correct lane position & L1, L2, L3, L4 \\
    H2 & Vehicle does not keep a safe distance to other road users & L1, L2, L4 \\
    H3 & Vehicle executes an unreasonable driving trajectory & L3, L4 \\
    H4 & Vehicle operates outside its defined operational design domain (ODD) & L1, L2, L3, L4 \\
    H5 & Vehicle fails to correctly respond to traffic signals & L1, L2, L3, L4 \\
    H6 & Failure to safely transfer control upon system degradation & L1, L2, L4 \\
    \bottomrule
  \end{tabularx}
\end{table}

\subsection{Model Control Structure}
\begin{figure*}[!t]
    \centering
    \includegraphics[width=\textwidth,keepaspectratio]{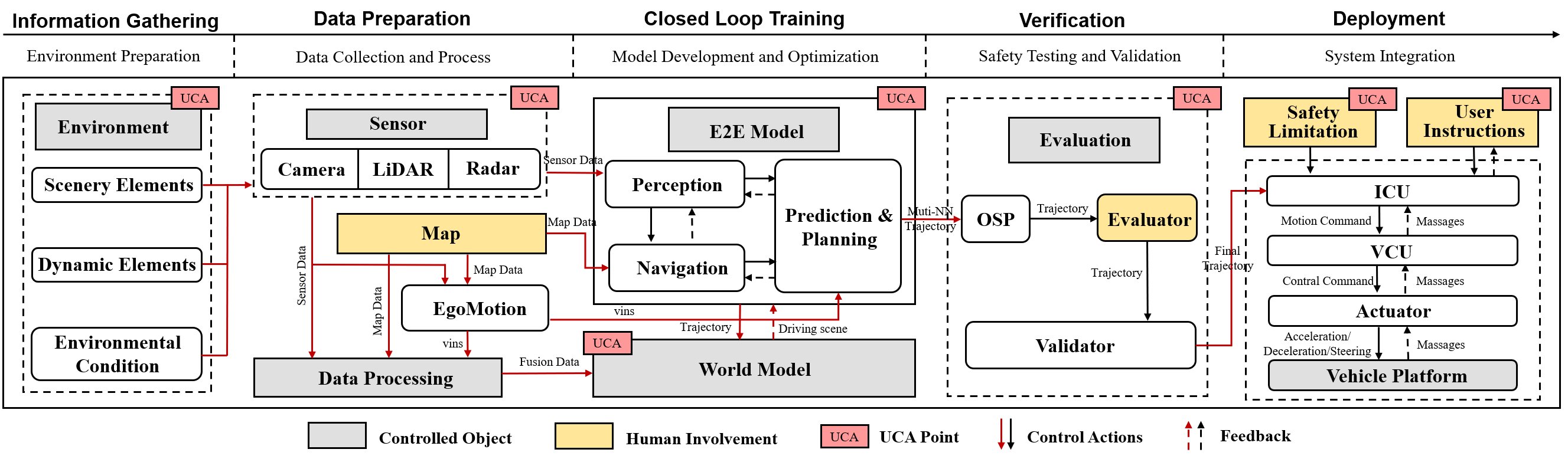}
    \caption{UniSTPA control loop structures applied on End-to-end Autonomous driving system. The entire diagram shows the stages of system development and deployment from left to right, as well as the key modules involved. ‘Gray boxes’ denotes functional modules of the technical system; ‘Yellow boxes’ denotes human involvement (such as developers and drivers); ‘UCA’ denotes potential unsafe control actions; ‘$\rightarrow$’ denotes control action, while ‘$\dashrightarrow$’ denotes feedback information; ‘red lines interactions between different stages; ‘black lines’ denotes interactions within the same stage; ‘ICU’ denotes interface control unit; ‘ICU’ denotes vehicle control unit.}
    \label{fig: UniSTPA}
\end{figure*}

The system control structure for this case is shown in \hyperref[fig: UniSTPA]{Fig 3}. This framework spans five key stages—information gathering, data preparation, closed-loop training, verification, and deployment—thereby constituting a complete E2E ADS development lifecycle. UniSTPA introduces innovative extensions along two dimensions: it expands the analysis scope from a single operational phase to the full lifecycle of an end-to-end autonomous driving system—including information gathering, data preparation, closed-loop training, verification, and deployment—and it transcends the “black-box” limitation by conducting fine-grained analysis of the model’s internal structure to identify safety risks within and across modules. 

\subsubsection{Information Gathering}
This stage focuses on environment preparation, including the modeling and configuration of road networks, scene elements, dynamic actors, and environmental conditions. The development team determines the required scenario coverage for model training, specifically designing highway–specific conditions such as ramp merges, diverges, and interwoven traffic flows. The resulting environment and scene elements serve as the basis for subsequent data acquisition and simulation, thereby establishing the preconditions for model training.

\subsubsection{Data Preparation}
This stage encompasses the acquisition and processing of sensor data. It involves the sensor suite (cameras, LiDAR, millimeter-wave radar), a high-definition map database, and a data processing module. The sensor suite captures raw environmental data, the map module supplies precise roadway information, and the EgoMotion module provides vehicle motion state. These data streams are fused and preprocessed by the data processing module to produce a unified world‐model input for E2E model training and decision making. During early development, this process may be conducted offline using recorded scenarios; after deployment, it transitions to an online, real‐time perception pipeline.

\subsubsection{Closed Loop Training}
This is the core of the E2E-based NOA system development, encompassing the design and optimization of the end-to-end model. This stage comprises three key submodules: \emph{Perception}, \emph{Navigation}, and \emph{Prediction \& Planning (PnP)} , which collectively define the functional architecture of the end-to-end model. The World Model component acts as an environment simulator, applying vehicle actions to the simulated environment and returning updated state information, thus creating a closed‐loop training process.

\subsubsection{Verification Stage}
This stage addresses safety testing and performance evaluation. It includes three principal components: Open Space Planning (OSP), an evaluator, and a validator. The evaluator assesses trajectories generated by the OSP, while the validator conducts comprehensive safety checks on the final trajectories. Any output that does not meet the defined safety criteria is rejected or fed back to the training stage for improvement. During this phase, the development team also refines the model or augments the test scenario library based on validation outcomes.

\subsubsection{Deployment Stage}
Finally, the validated model is integrated into the production vehicle platform. The control structure at this stage aligns with a conventional vehicle control loop. During on-road operation, the end-to-end model functions as the controller, issuing commands to steering and braking actuators, while vehicle state and environmental feedback are continuously fed back to the perception module, thereby maintaining a closed‐loop control system.

\subsection{Identify UCAs}
Table \ref{tab:ucas} lists the identified representative UCAs. Traditional STPA identifies UCAs into four categories: \textit{not provided}, \textit{provided improperly}, \textit{mistimed provision}, and \textit{inappropriate duration}—primarily applied during the system’s operational phase. In the UniSTPA framework, we extend this classification and screening process to every phase of the E2E‐based NOA system lifecycle. Owing to space limitations, we do not enumerate all possible UCAs. Instead, we select a subset of representative key UCAs to illustrate UniSTPA’s broader coverage compared to the traditional approach.

As shown in Table \ref{tab:ucas}, UniSTPA identifies UCAs that are difficult to uncover using traditional STPA, particularly in the Information Collection, Data Preparation, and Closed‐Loop Training stages. For example, UCA‐IG1 and UCA‐IG2 reflect how incomplete environmental scene design can induce biases in training data; UCA‐DP1 through UCA‐DP3 reveal how deficiencies in the data preparation stage compromise system safety; and UCA‐LT1 through UCA‐LT3 demonstrate potential issues arising within the training processes of the end‐to‐end model’s internal modules. Had one employed only the traditional STPA focus on system operation, the majority of these UCAs would remain undetected, thereby underscoring the necessity and effectiveness of UniSTPA for safety analysis in end‐to‐end autonomous driving systems.

\subsection{Identify Causal Scenarios}
Based on the identified UCAs, this section further analyzes the underlying causal scenarios that lead to these unsafe control actions. In the UniSTPA framework, causal analysis not only considers the controller, control logic, process models, and feedback factors emphasized in traditional STPA but also highlights the internal structure unique to end-to-end models and the lifecycle-wide causality. Table \ref{tab:cs} presents the causal scenario analysis for several representative UCAs.

\begin{table*}[htpb]
  \caption{Representative UCAs and Associated System Hazards. ‘IG’ denotes Information Gathering; ‘DP’ denotes Data Preparation; ‘LT’ denotes Closed‐Loop Training; ‘VF’ denotes Validation; and ‘DT’ denotes Deployment.}
  \label{tab:ucas}
  \centering
  \begin{tabularx}{\textwidth}{@{}l l X l l@{}}
    \toprule
    \textbf{UCA ID} & \textbf{Control Action} & \textbf{Description of Unsafe Control Action} & \textbf{Failure Mode} & \textbf{Hazards} \\
    \midrule
    UCA-IG1   & Environment scene element configuration
               & Failure to include complex vehicle weaving scenarios at highway ramp merges and diverges
               & Not Provided
               & H2, H3 \\
    UCA-IG2   & Environmental boundary configuration
               & Insufficient lighting coverage, omitting high-contrast tunnel entrance/exit scenarios
               & Provided Improperly
               & H1, H4 \\
    \midrule
    UCA-DP1   & Sensor data acquisition
               & Inadequate sensor data under adverse weather at high speeds
               & Not Provided
               & H4 \\
    UCA-DP2   & Map data update
               & Incomplete marking of temporary speed limits in highway construction zones
               & Provided Improperly
               & H5 \\
    UCA-DP3   & Data processing and fusion
               & Failure to resolve conflicting sensor data under high-speed conditions
               & Provided Improperly
               & H1, H2 \\
    \midrule
    UCA-LT1   & Perception module training
               & Insufficient detection of small obstacles (debris, cargo) on highway surfaces
               & Provided Improperly
               & H1, H2 \\
    UCA-LT2   & Navigation module update
               & Incorrect interpretation of navigation instructions in complex interchanges
               & Provided Improperly
               & H3 \\
    UCA-LT3   & Prediction \& planning module training
               & Inaccurate prediction of other vehicles’ lane-change intentions
               & Mistimed Provision
               & H2, H3 \\
    UCA-LT4   & End-to-end model optimization
               & Overemphasis on high-speed cruise efficiency at the expense of ramp merge safety
               & Provided Improperly
               & H1, H2, H3 \\
    \midrule
    UCA-VF1   & Trajectory validation
               & Validator fails to identify trajectory conflicts in dense high-speed traffic
               & Not Provided
               & H2, H3 \\
    UCA-VF2   & OSP evaluation
               & Evaluator assigns excessively high scores to emergency‐maneuver trajectories
               & Provided Improperly
               & H2, H3 \\
    \midrule
    UCA-DT1   & Safety limitation configuration
               & Safety thresholds set too leniently, permitting operation under unsuitable conditions
               & Provided Improperly
               & H4 \\
    UCA-DT2   & Control command execution
               & Excessive delay in executing high-speed emergency avoidance commands
               & Mistimed Provision
               & H2 \\
    UCA-DT3   & Driver takeover prompt
               & Failure to prompt the driver for takeover upon detecting ODD boundary breaches
               & Not Provided
               & H4, H6 \\
    \bottomrule
  \end{tabularx}
\end{table*}

\begin{table*}[!ht]
  \caption{Causal Scenario Analysis for Representative UCAs}
  \label{tab:cs}
  \centering
  \begin{tabularx}{\textwidth}{@{}l l c X@{}}
    \toprule
    \textbf{UCA ID} & \textbf{CS ID} & \textbf{Stage} & \textbf{Causal Scenario Description} \\
    \midrule
    UCA-IG1 & CS-IG1-1 & 
      & Insufficient recognition of vehicle weaving behavior characteristics at highway ramp areas. \\
           & CS-IG1-2 & Information Gathering
      & Failure to account for regional variations in ramp design affecting driving behavior. \\
           & CS-IG1-3 & 
      & Inadequate modeling of multi-vehicle interactions under high-density traffic conditions. \\
    \midrule
    UCA-DP3 & CS-DP3-1 & 
      & Improper single-source prioritization strategy during multi-sensor data conflicts. \\
           & CS-DP3-2 & Data Preparation
      & Sensor calibration biases leading to inaccuracies in data fusion. \\
           & CS-DP3-3 & 
      & Absence of adaptive data-processing algorithms for varying environmental conditions. \\
    \midrule
    UCA-LT1 & CS-LT1-1 & 
      & Insufficient sample volume of small obstacles (e.g., debris) in highway training data. \\
           & CS-LT1-2 & Closed Loop Training
      & Deficiencies in annotation quality for small targets in high-speed scenarios. \\
           & CS-LT1-3 & 
      & Perception network architecture limitations in extracting features of dynamic small targets. \\
    \midrule
    UCA-LT3 & CS-LT3-1 & 
      & Inadequate coverage of complex lane-change scenarios in training data. \\
           & CS-LT3-2 & 
      & Simplified model structure that fails to capture subtle precursors of lane-change intentions. \\
           & CS-LT3-3 & Closed Loop Training
      & Information transmission delays between modules resulting in outdated predictions. \\
           & CS-LT3-4 & 
      & Loss-function weightings that insufficiently emphasize accuracy in lane-change prediction. \\
    \midrule
    UCA-VF1 & CS-VF1-1 & 
      & Validation scenario library lacks coverage of complex high-speed traffic scenarios. \\
           & CS-VF1-2 & Verification
      & Insufficient sensitivity to trajectory conflicts arising from multi-vehicle interactions. \\
           & CS-VF1-3 & 
      & Overly simplified risk assessment model. \\
    \midrule
    UCA-DT3 & CS-DT3-1 & 
      & ODD-monitoring system lacks sensitivity to sudden environmental changes. \\
           & CS-DT3-2 & Deployment
      & Excessive communication latency between system monitoring and human–machine interaction modules. \\
           & CS-DT3-3 & 
      & Inappropriately configured takeover decision thresholds. \\
           & CS-DT3-4 & 
      & Insufficient salience of takeover prompts under high-speed conditions. \\
    \bottomrule
  \end{tabularx}
\end{table*}

\begin{table*}[!t]
  \caption{Safety Requirements Based on Causal Scenarios}
  \label{tab:sr}
  \centering
  \begin{tabularx}{\textwidth}{@{}l l X@{}}
    \toprule
    \textbf{Causal Scenario ID} & \textbf{Requirement ID} & \textbf{Safety Requirement Description} \\
    \midrule
    CS-IG1-1, CS-IG1-2 & SR-IG1-1
      & Establish a comprehensive scenario library and training guidelines for multi-regional highway ramp merging and weaving behaviors. \\
    CS-IG1-3 & SR-IG1-2
      & Develop modeling tools for complex multi-vehicle interaction traffic flows. \\
    \midrule
    CS-DP3-1 & SR-DP3-1
      & Implement a Bayesian-based dynamic sensor reliability assessment mechanism. \\
    CS-DP3-2 & SR-DP3-2
      & Establish an automatic sensor self-calibration and anomaly detection system for high-speed operation. \\
    CS-DP3-3 & SR-DP3-3
      & Develop an environment-adaptive data weighting strategy for robust data fusion. \\
    \midrule
    CS-LT1-1, CS-LT1-2 & SR-LT1-1
      & Construct a dedicated dataset and annotation standards for small obstacle detection on high-speed roads. \\
    CS-LT1-3 & SR-LT1-2
      & Optimize the network architecture to enable multi-scale feature extraction for small dynamic targets. \\
    \midrule
    CS-LT3-1 & SR-LT3-1
      & Augment the training dataset with complex lane-change scenarios and apply data augmentation techniques. \\
    CS-LT3-2 & SR-LT3-2
      & Incorporate temporal attention mechanisms to enhance recognition of lane-change precursors. \\
    CS-LT3-3 & SR-LT3-3
      & Implement low-latency, feature-level information sharing mechanisms between modules. \\
    CS-LT3-4 & SR-LT3-4
      & Redesign the loss function to optimize accuracy in high-speed lane-change intention prediction. \\
    \midrule
    CS-VF1-1 & SR-VF1-1
      & Build a validation scenario library for critical high-speed traffic conditions and establish coverage evaluation criteria. \\
    CS-VF1-2, CS-VF1-3 & SR-VF1-2
      & Implement a reinforcement learning-based trajectory risk assessment algorithm. \\
    \midrule
    CS-DT3-1 & SR-DT3-1
      & Deploy a high-sensitivity ODD boundary monitoring and environmental trend prediction system. \\
    CS-DT3-2 & SR-DT3-2
      & Design a low-latency transmission and quality monitoring architecture for safety-critical information. \\
    CS-DT3-3 & SR-DT3-3
      & Establish a multi-level dynamic takeover decision mechanism based on ODD margin. \\
    CS-DT3-4 & SR-DT3-4
      & Define design standards for multimodal HMI takeover interfaces under high-speed conditions. \\
    \bottomrule
  \end{tabularx}
\end{table*}

From the causal scenario analysis in Table \ref{tab:cs}, it can be observed that UniSTPA thoroughly examines potential issues across all stages of the end-to-end model lifecycle. For example, CS-LT1-1 through CS-LT1-3 reveal data quality, annotation, and network architecture issues within perception module training; CS-LT3-1 through CS-LT3-3 analyze multidimensional factors in the PnP module, including data integrity, model structure and module interaction; And CS-DT3-1 through CS-DT3-4 examine causes of delayed takeover prompts from the perspectives of operational design domain (ODD) monitoring, communication latency, decision thresholds, and human–machine interaction. Such multi-stage, multi-level causal analysis is not readily achievable with traditional STPA methods.

Notably, UniSTPA not only addresses issues internal to individual modules (e.g., CS-LT1-3 concerning suboptimal network design) but also emphasizes systemic risks arising from inter-module interactions (e.g., CS-LT3-3 regarding information transmission delays between modules). This analytical perspective better aligns with the holistic nature of end-to-end autonomous driving systems.

\subsection{Derive Safety Requirements}
Based on the identified UCAs and causal scenarios, this section derives the corresponding safety requirements for the E2E-based NOA system. These requirements encompass not only control strategies during system operation but also safety measures across all lifecycle stages. Table \ref{tab:sr} presents the safety requirements associated with representative causal scenarios.

\begin{figure*}[!t]
    \centering
    \includegraphics[width=\textwidth,keepaspectratio]{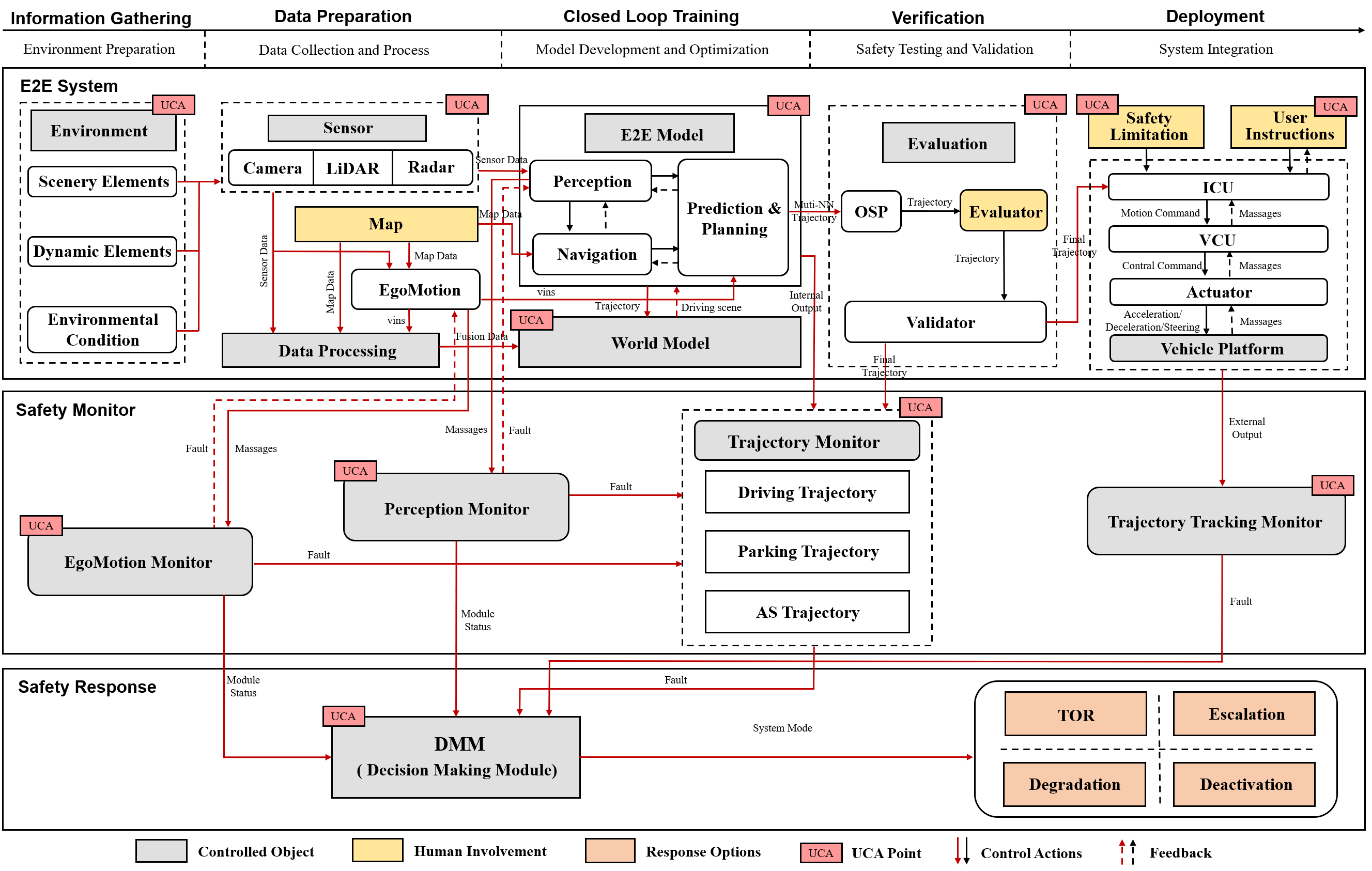}
    \caption{UniSTPA Control Loop with Safety Monitor and Safety Response. ‘AS Trajectory’ denotes Active-Safety Trajectory; ‘TOR’ denotes Take-Over Request.}
    \label{fig: UniSTPA-S}
\end{figure*}

To ensure effective implementation of these requirements and to prevent the occurrence of UCAs, we have designed a safety monitoring and response mechanism, as shown in \hyperref[fig: UniSTPA-S]{Fig 4}. This mechanism comprises two core subsystems: \textit{Safety Monitor} and \textit{Safety Response}. The Safety Monitoring subsystem consists of the EgoMotion Monitor, the Perception Monitor, and the Trajectory Monitor, which are responsible for continuously assessing vehicle motion state, environmental perception quality, and trajectory rationality, respectively. By interacting in real time with the data processing, perception, and trajectory planning modules, these monitors evaluate system status and identify potential risks.

The Safety Response subsystem implements a tiered response strategy with four levels: Takeover Request (TOR), Performance Degradation, Functional Escalation, and System Deactivation. Depending on the severity and type of risk detected, the Decision-Making Module (DMM) selects the appropriate response and issues commands to the actuators via the control interface.

This lifecycle-spanning safety monitoring and response mechanism not only addresses immediate operational risks but also enhances safety during development through closed-loop feedback. For example:  
\textit{i)} Issues detected by the Trajectory Monitor can be fed back to the validation stage to refine the scenario library;  
\textit{ii)} Defects identified by the Perception Monitor can inform improvements to model architecture and training strategies in the closed-loop training stage;  
\textit{iii)} Anomalies flagged by the EgoMotion Monitor can guide sensor calibration and fusion algorithm optimization during the data preparation stage.

Through this closed‐loop feedback mechanism, the safety monitoring and response system continuously enhances the safety performance of the end‐to‐end autonomous driving system across all lifecycle stages.

\section{CONCLUSION AND FUTURE WORK}\label{EE}

This paper introduces the Unified System Theoretic Process Analysis (UniSTPA) framework, which provides a systematic solution to the key challenges of safety analysis for modular E2E autonomous driving systems.
Traditional safety analysis methods face multiple limitations when applied to E2E models, including insufficient coverage of the entire lifecycle due to analysis of only a single stage and difficulty in analyzing internal hazards in learning-based systems. UniSTPA overcomes these limitations through a dual expansion along both the \textit{lifecycle} dimension and the \textit{structural‐depth} dimension. Moreover, the framework incorporates a closed‐loop safety monitoring and response mechanism that enables continuous improvement from risk identification to system optimization. A case study on a highway NOA function validates UniSTPA’s effectiveness and advantages, successfully uncovering multi‐stage hazards—such as scene design defects, sensor fusion issues, and specific internal model flaws—that traditional methods would overlook. Through multi‐level causal analysis, UniSTPA traces these hazards to deeper issues in data quality, model architecture, and optimization objectives, and subsequently derives a comprehensive set of safety requirements spanning the entire lifecycle.

Future research will extend the UniSTPA framework along three key directions: \textit{i)} Develop an integrated analysis toolchain to automate UniSTPA workflows and reduce complexity and implementation cost; \textit{ii)} Establish quantitative analysis methods for end‐to‐end models that combine formal verification techniques with statistical guarantees to provide measurable safety indices and mathematical assurances for deep‐learning models; \textit{iii)} Align UniSTPA processes with emerging AI safety standards (e.g., ISO/PAS 8800) to create a regulatory‐compliant safety‐analysis workflow. These research efforts will collectively advance the safe and reliable development of end‐to‐end autonomous driving systems and drive UniSTPA from a theoretical framework to practical.

\bibliographystyle{IEEEtran}
\bibliography{arxiv}

\end{document}